\documentclass[10pt,twocolumn,letterpaper]{article}

\usepackage[pagenumbers]{cvpr} 

\usepackage{graphicx}
\usepackage{amsmath}
\usepackage{amssymb}
\usepackage{booktabs}
\usepackage{makecell}
\usepackage{multirow}
\usepackage{colortbl}
\usepackage{tabu}
\usepackage{pifont}
\newcommand{\cmark}{\ding{51}\xspace}%
\newcommand{\xmarkg}{\textcolor{lightgray}{\ding{55}}\xspace}%
\usepackage[misc]{ifsym}
\usepackage{flushend}

\usepackage[pagebackref,breaklinks,colorlinks]{hyperref}

\usepackage[capitalize]{cleveref}
\crefname{section}{Sec.}{Secs.}
\Crefname{section}{Section}{Sections}
\Crefname{table}{Table}{Tables}
\crefname{table}{Tab.}{Tabs.}

\newcommand{\Rmnum}[1]{\expandafter\@slowromancap\romannumeral #1@}

\usepackage{color}
\usepackage{xcolor}

\newcommand{\ntacc}{N-acc.\xspace}

\let\oldsubsection\subsection
\renewcommand{\subsection}[1]{\oldsubsection{#1} }

\def\cvprPaperID{***} 
\def\confName{CVPR}
\def\confYear{2023}

\begin{document}

\title{GREC: Generalized Referring Expression Comprehension}
\author{
Shuting He\footnotemark[2]
\qquad
Henghui Ding\footnotemark[2]~~$^{\textrm{\Letter}}$
\qquad
Chang Liu
\qquad
Xudong Jiang\\
Nanyang Technological University\\
\href{https://henghuiding.github.io/GRES}{https://henghuiding.github.io/GRES}
}
\vspace{-5mm}

\twocolumn[{%
\renewcommand\twocolumn[1][]{#1}%
\maketitle 
\begin{center} 
\centering 
\includegraphics[width=1\textwidth]{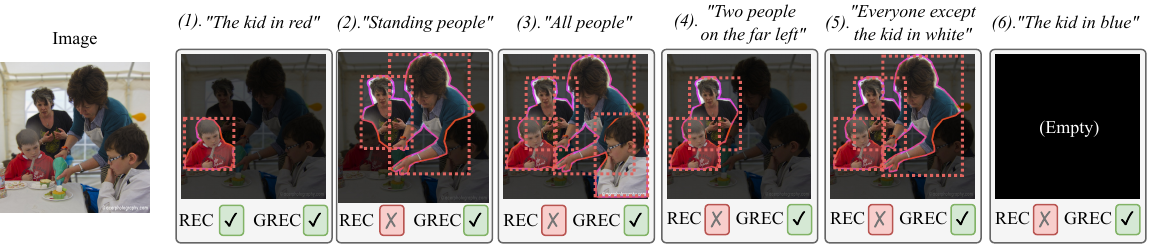} 
\vspace{-6.6mm}
\captionof{figure}{Classic Referring Expression Comprehension (REC) only supports expressions that indicate a single target object, \eg, (1). Compared with classic REC, the proposed \textbf{Generalized Referring Expression Comprehension (GREC)} supports expressions indicating an
\textbf{\textit{arbitrary number}} of target objects, for example, multi-target expressions like ({2})-({5}), and no-target expressions like ({6}).}
\label{fig:fig1} 
\vspace{6mm}
\end{center}%
}]

\renewcommand{\thefootnote}{\fnsymbol{footnote}}
\footnotetext[2]{Equal contribution.}
\footnotetext[0]{${\textrm{\Letter}}$ henghui.ding@gmail.com.}

\begin{abstract}
  The objective of Classic Referring Expression Comprehension (REC) is to produce a bounding box corresponding to the object mentioned in a given textual description. Commonly, existing datasets and techniques in classic REC are tailored for expressions that pertain to a single target, meaning a sole expression is linked to one specific object. Expressions that refer to multiple targets or involve no specific target have not been taken into account. This constraint hinders the practical applicability of REC. This study introduces a new benchmark termed as Generalized Referring Expression Comprehension (GREC). This benchmark extends the classic REC by permitting expressions to describe any number of target objects. To achieve this goal, we have built the first large-scale GREC dataset named gRefCOCO. This dataset encompasses a range of expressions: those referring to multiple targets, expressions with no specific target, and the single-target expressions. The design of GREC and gRefCOCO ensures smooth compatibility with classic REC. The proposed gRefCOCO dataset, a GREC method implementation code, and GREC evaluation code are available at \href{https://github.com/henghuiding/gRefCOCO}{https://github.com/henghuiding/gRefCOCO}.

\end{abstract}

\section{Introduction}
\label{sec:intro}

The fundamental objective of REC is to pinpoint the specified target object and subsequently delineate its location using a bounding box. The potential of REC extends across a spectrum of applications, encompassing domains like video production, human-machine interaction, and robotics. Presently, the predominant methodologies adhere to REC guidelines as stipulated in prominent datasets such as  ReferIt \cite{kazemzadeh-etal-2014-referitgame} and RefCOCO \cite{yu2016modeling,mao2016generation}. These approaches have demonstrated substantial advancements in recent times~\cite{hu2016natural,li2023transformer,wang2019neighbourhood,liu2019learning,yang2019fast,zhuang2018parallel,yang2020improving,liao2020real,RefCLIP,hu2017modeling,zhang2018grounding,hong2019learning,chen2018real,sun2022proposal, deng2021transvg,MDETR,GroundingDINO}.

In recent research by GRES~\cite{GRES}, the scope of classic Referring Expression Segmentation (RES)~\cite{VLTPAMI,ding2021vision,phraseclick,liu2022instance,liu2023multi,wu2022towards} is broadened to Generalized Referring Expression Segmentation (GRES). Furthermore, in the video domain, MeViS~\cite{MeViS}, sourcing videos mainly from MOSE~\cite{MOSE}, builds a large-scale motion expression video segmentation dataset that supports expressions referring to a varying number of object(s) in the given video. As a follow-up to GRES~\cite{GRES}, this study investigates the constraints inherent in classic Referring Expression Comprehension (REC). Furthermore, a novel benchmark named Generalized Referring Expression Comprehension (GREC) is introduced in this work as an extension of the classic REC.

\textbf{Limitations of classic REC.} Most classic REC methods have some strong pre-defined constraints to the task. Firstly, classic REC disregards no-target expressions, \ie, expressions that fail to correspond with any object within the image. Consequently, if the designated target isn't present in the input image, the behavior of existing REC methodologies becomes undefined. In practical contexts, adhering to this constraint mandates that the input expression aligns with an image object; otherwise, complications invariably arise. Secondly, most existing datasets, \eg, the most popular RefCOCO \cite{yu2016modeling,mao2016generation}, lack provisions for multi-target expressions, which refer to multiple objects within the given image. Consequently, searching for objects necessitates multiple inputs, effectively mandating a one-by-one object retrieval process. For example, as shown in \cref{fig:fig1}, grounding \textit{``All people''} necessitates four distinct expressions, thus entailing four instances of model invocation. Our experimentation highlights the inadequacy of classic REC techniques trained on existing datasets to adapt effectively to these scenarios.

\begin{table}[t]
  \renewcommand\arraystretch{1.2}
  \centering
  \footnotesize
  \caption{Comparison among different referring expression comprehension datasets, including ReferIt\cite{kazemzadeh-etal-2014-referitgame}, RefCOCO(g)\cite{yu2016modeling,mao2016generation}, PhraseCut\cite{wu2020phrasecut}, and our proposed \textbf{gRefCOCO}. Multi-target: expression that specifies multiple objects in the image. No-target: expression that does not touch on any object in the image.}\vspace{-1.6mm}
  \setlength{\tabcolsep}{1.36mm}{\begin{tabular}{lcccc}
    \specialrule{.1em}{.05em}{.05em} 
          & ReferIt & RefCOCO(g)& PhraseCut&  \textbf{gRefCOCO}\\
          \cline{2-5}
          Image Source  & CLEF\cite{grubinger2006iapr} & COCO\cite{lin2014microsoft} & VG\cite{krishna2017visual}  & COCO\cite{lin2014microsoft} \\
          Multi-target & {\xmarkg} & {\xmarkg} & {(fallback)} & {\cmark} \\
          No-target & {\xmarkg} & {\xmarkg} & {\xmarkg}  & {\cmark} \\
          \makecell[c]{Expression type} & free  & free  & templated & free \\
    \specialrule{.1em}{.05em}{.05em} 
    \end{tabular}}%
  \label{tab:dataset_compare}%
\end{table}%

\begin{figure*}[t]
  \centering
  \vspace{0.35em}
  \hfill
  \begin{subfigure}[t]{0.5\linewidth}
      \centering
      \includegraphics[width=\textwidth]{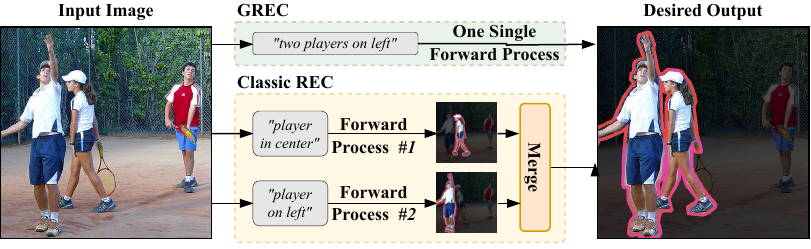}
      \caption{Multi-target: Selecting multiple objects in one single forward process.}
      \label{fig:app_mt}
  \end{subfigure}
  \hfill
  \begin{subfigure}[t]{0.47\linewidth}
      \centering
      \includegraphics[width=\textwidth]{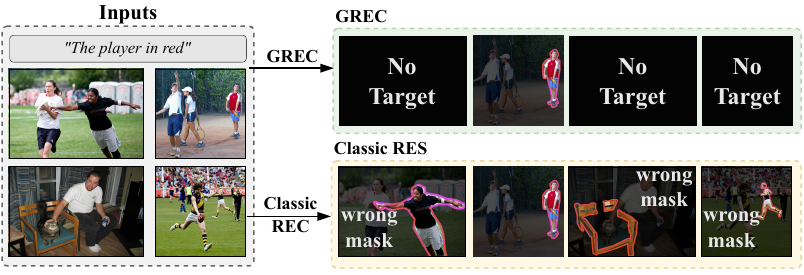}
      \caption{No-target: Retrieving images that contain the object.}
      \label{fig:app_nt}
  \end{subfigure}
  \hfill
  \caption{More applications of GREC brought by supporting multi-target and no-target expressions compared to classic REC.}
  \label{fig:gres_app}
\end{figure*}

\textbf{New benchmark and dataset.} In this work, we propose a new benchmark, termed as Generalized Referring Expression Comprehension (GREC). This benchmark diverges from classic REC setups by accommodating expressions that can pertain to any quantity of target objects. Similar to the classic REC paradigm, GREC operates by taking both an image and a referring expression as input. In contrast to classic REC, as shown in \cref{fig:fig1}, GREC presents an advancement by incorporating additional capabilities. Specifically, GREC extends its support to encompass multi-target expressions, which entail the specification of multiple target objects through a single expression, \eg, \textit{``Everyone except the kid in white''}. Furthermore, GREC accommodates no-target expressions that don't reference any object in the image. An example of this would be \textit{``the kid in blue''}. 
This enhancement introduces a significantly higher level of versatility to input expressions, thereby augmenting the utility and resilience of Referring Expression Comprehension in practical applications. However, it's noteworthy that existing referring expression datasets~\cite{kazemzadeh-etal-2014-referitgame,yu2016modeling,mao2016generation} do not contain multi-target expression nor no-target samples, but exclusively encompass single-target expression samples, as shown in \cref{tab:dataset_compare}. To support research efforts in the realm of practical referring comprehension, we have built a new dataset for GREC, named gRefCOCO. This dataset serves as a valuable extension to RefCOCO and encompasses two distinct categories of samples: multi-target samples, in which the expression refers to two or more targets within the input image, and no-target samples, in which the expression fails to match any object in the input image.

We establish new evaluation metrics tailored for GREC and conduct extensive experiments. The outcomes of these experiments unequivocally indicate that prevailing state-of-the-art REC techniques exhibit limitations in effectively addressing the newly introduced GREC task. A primary contributing factor to this outcome is the inherent assumption within these methods that invariably presupposes the existence of a solitary target object, thereby leading to the output of \underline{\textbf{\textit{a singular bounding box}}}. However, in the context of GREC, the number of the target objects is unlimited and hence the number of desired bounding boxes ranges from 0 to many, rather than ``1'' as assumed in REC.

\section{Task Setting and Dataset}

\subsection{GREC Settings}

\textbf{Revisit of REC.} In the classic Referring Expression Comprehension (REC) paradigm, the process involves inputting an image and a corresponding expression. The ultimate goal is to generate \textbf{a bounding box} delineating the object in the image to which the expression refers. The existing REC approach overlooks the possibility of no-target expressions, and the datasets currently available exclusively consist of single-target expressions. Consequently, prevailing models are susceptible to generating erroneous instances when confronted with input expressions that either pertain to multiple targets or nothing at all within the given image.

\textbf{Generalized REC.} To address these limitations in classic REC, we propose a benchmark called Generalized Referring Expression Comprehension (GREC) that allows expressions indicating arbitrary number of target objects. In contrast to classic REC, which outputs a single bounding box for a referring expression, GREC aims to output a set of bounding boxes $B={b_i}$, where each $b_i$ corresponds to an object among all the target objects referred by the given expression. The number of bounding boxes can range from 0 to multiple, depending on the given expression. If the expression does not refer to any object in the image, \ie, no-target expression, then no bounding box should be output.

\textbf{The significance and benefits of extension.} The applications of multi-target and no-target expressions encompass more than just identifying multiple targets and disregarding inappropriate expressions that do not match any object. These extensions also play a significant role in imbuing referring expression comprehension with heightened realism and enabling advanced use cases. For example, leveraging the capabilities of multi-target expressions introduces innovative possibilities. Expressions such as ``\textit{all people}'' and ``\textit{two players on left}'' can be utilized as inputs to simultaneously identify multiple objects within a single forward process (see \cref{fig:app_mt}). Additionally, expressions like ``\textit{foreground}'' and ``\textit{kids}'' can be employed to achieve user-defined open vocabulary perception~\cite{wu2023towards,han2023global,wu2023betrayed,zhanghui2021,D2Zero,PADing}, adding a layer of flexibility to the process.
Incorporating no-target expressions into the framework enables users to apply identical expressions across a collection of images, facilitating the identification of images that feature the object(s) mentioned in the language expression, as shown in \cref{fig:app_nt}. This functionality proves particularly valuable when users aim to locate and extract specific elements from a group of images, akin to image retrieval, albeit with greater precision and adaptability.
Furthermore, the incorporation of both multi-target and no-target expressions enhances the model's reliability and resilience in the face of realistic scenarios where diverse types of expressions can unexpectedly emerge. This takes into account the fact that users might inadvertently or deliberately introduce errors in their sentences, highlighting the importance of accommodating a wide range of potential inputs.

\subsection{gRefCOCO: A Large-scale GREC Dataset}

Please kindly refer to GRES~\cite{GRES} for more details about the proposed gRefCOCO dataset.

\section{Experiments and Discussion}

\subsection{Evaluation Metrics} \label{sec:metrics}

Different from GRES~\cite{GRES}, GREC task demands that methods generate precise bounding boxes for each individual instance of the referred targets within an image. In essence, these methods should exhibit the capacity to effectively differentiate between different instances. This requirement holds significance and is crucial for GREC, given that the desired outputs are bounding boxes. It ensures that the achieved outcomes align closely with the intended objective. Otherwise, there's a risk of yielding erroneous outcomes, such as predicting a single oversized bounding box that covers the entire image.

Each sample in classic REC has only one ground truth bounding box and one predicted bounding box, thus the prediction can be regarded as either a true positive (TP) or a false positive (FP). Previous classic REC methods adopt Precision@0.5 (\textit{a.k.a} top-1 accuracy) as the metric, where a prediction is considered TP if its IoU with ground truth bounding box is greater than 0.5. However, since a GREC sample has an unlimited number of ground truth bounding boxes and an unlimited number of predicted bounding boxes, the way of determining TP by IoU does not reflect the quality of prediction. We set two new metrics for GREC: Precision@(F$_1$=1, IoU$\ge$0.5) and No-target accuracy (\ntacc). 

\textbf{Precision@(F$_1$=1, IoU$\ge$0.5)} computes the percentage of samples that have the F$_1$ score of 1 with the IoU threshold set to 0.5. Given a sample and its predicted/ground-truth bounding boxes, a predicted bounding box is regarded as a TP if it has a matched (IoU$\geq$0.5) ground-truth bounding box. If there are several predicted bounding boxes matching one ground-truth bounding box, only the one with the highest IoU is TP while others are FP. The ground-truth bounding boxes having no matched bounding box are FN while the predicted bounding boxes having no matched ground-truth are FP. We measure the F$_1$ score of this sample by $F_1=\frac{2TP}{2TP+FN+FP}$. If the F$_1$ score is 1.0, this sample is considered successfully predicted. As for no-target samples, the F$_1$ score is regarded as 1 if there is no predicted bounding box otherwise 0. Then we compute the ratio of such successfully predicted samples, \ie, Precision@(F$_1$=1, IoU$\ge$0.5).

\textbf{\ntacc}~(no-target accuracy) assesses the model's proficiency in no-target identification. For a no-target sample, prediction without any bounding box is considered a true positive (TP), otherwise false negative (FN). Then, \ntacc~measures the model's performance on identifying no-target samples: \ntacc~= $\frac{\mathit{TP}}{\mathit{TP}+\mathit{FN}}$.

Following an extensive array of experiments, we earnestly recommend that future research endeavors prioritize the adoption of ``\textbf{Precision@(F$_1$=1, IoU$\ge$0.5)}'' as the principal metric. This choice is rooted in its superior capacity to accurately capture the quality of grounding within the GREC context. The evaluation codes can be found from \href{https://github.com/henghuiding/gRefCOCO}{https://github.com/henghuiding/gRefCOCO}, where we also provide a GREC implementation code of MDETR~\cite{MDETR}.

\begin{table}[t]
    \renewcommand\arraystretch{1.05}
    \centering
    \small
    \caption{Ablation study on how to select bounding boxes for multi-target and no-target expressions.}
    \vspace{-1.6mm}
    \centering
       \setlength{\tabcolsep}{3mm}{\begin{tabular}{l|ccc}
        \specialrule{.1em}{.05em}{.05em} 
        Strategy& Pr@(F$_1$=1, IoU$\ge$0.5)   & AP  & \ntacc   \\
        \hline\hline
        Top-1   & 0.0 & 26.2 & 0.0 \\
        Top-5   & 0.0& 52.8  & 0.0 \\
        Top-10  & 0.0 & 53.3  & 0.0  \\
        Top-100  & 0.0 & 53.5  & 0.0  \\
        Threshold-0.5  & 37.0  &  52.6  & 32.2   \\
        Threshold-0.6  &  38.9 & 52.5   &33.9   \\
        Threshold-0.7  & 41.5 & 52.3  & 36.1  \\
        Threshold-0.8  & 44.7  & 51.7   & 39.2  \\
        Threshold-0.9  &  51.2 & 50.8   & 45.7  \\
          \specialrule{.1em}{.05em}{.05em} 
  \end{tabular}}
\label{tab:selectingBox}
\end{table}

\begin{table*}[t]
  \renewcommand\arraystretch{1.05}
  \begin{center}
  \small
  \caption{GREC benchmark results on gRefCOCO dataset. $\dagger$ indicates that the detection head has been adapted to generate multiple bounding boxes and subsequently select the target box(es) using a threshold-based criterion. Threshold is set to 0.7 for all the methods.}\label{tab:results_grec}
  \vspace{-1.6mm}
  \centering
     \setlength{\tabcolsep}{0.96mm}{\begin{tabular}{l|c|c|cc|cc|cc}
      \specialrule{.1em}{.05em}{.05em} 
      \multirow{2}{*}{Methods} &\multirow{2}{*}{\shortstack{Visual\\Encoder}} & \multirow{2}{*}{\shortstack{Textual\\Encoder}}& \multicolumn{2}{c|}{val} & \multicolumn{2}{c|}{testA} & \multicolumn{2}{c}{testB} \\
             &  &  &Pr@(F$_1$=1, IoU$\ge$0.5)   & \ntacc   &Pr@(F$_1$=1, IoU$\ge$0.5)   &  \ntacc   &Pr@(F$_1$=1, IoU$\ge$0.5)    & \ntacc   \\
        \hline\hline
        MCN$^\dagger$~\cite{luo2020multi}   & DarkNet-53 &GRU& 28.0 &30.6  & 32.3 & 32.0 & 26.8 & 30.3 \\
        VLT$^\dagger$~\cite{ding2021vision}   & DarkNet-53 & GRU& 36.6 & 35.2  & 40.2 & 34.1 & 30.2 & 32.5 \\
        MDETR$^\dagger$~\cite{MDETR}  & ResNet-101 & RoBERTa& 42.7 & 36.3 & 50.0 & 34.5 & 36.5 & 31.0   \\
        UNINEXT$^\dagger$~\cite{UNINEXT} & ResNet-50 & BERT&58.2  &  50.6 & 46.4 & 49.3 & 42.9 & 48.2 \\
        \specialrule{.1em}{.05em}{.05em} 
\end{tabular}}
\end{center}
\vspace{-1.6mm}
\footnotesize{\textit{It is worth noting that the original UNINEXT model in \cite{UNINEXT} is pre-trained using images from val/testA/testB sets, which introduces a form of information leakage in the context of GREC/REC tasks. Here, we take a rigorous training setup and exclude the val/testA/testB images when training UNINEXT.}}
\end{table*}

\subsection{Results on GREC}
\textbf{Does the top-1 or top-k strategy used in classic REC work for multi-target and no-target expressions of GREC?} Existing state-of-the-art REC methods typically select the top-1 bounding box as the final output~\cite{MDETR}, or just predict a single bounding box~\cite{luo2020multi} as the output. When it comes to GREC task where the target objects varies from 0 to many, does the top-1 or top-k strategy still works? To answer this question and study the better prediciton way for GREC, we conduct an ablation study on the strategy of selecting output bounding boxes for multi-target and no-target samples in \cref{tab:selectingBox}, based on MDETR~\cite{MDETR} that predicts 100 bounding boxes. When selecting the top-1
bounding box, the Precision@(F$_1$=1, IoU$\ge$0.5) and \ntacc~are both 0. This rationale stems from the fact that the top-1 strategy predicts every sample to have only one object. Consequently, the highest achievable F$_1$ score for multi-target samples becomes $\frac{2}{2+1+0}=0.67$. In other words, there are two target objects, and one of them aligns accurately with the predicted bounding box, resulting in the best F$_1$ score of 0.67 for the top-1 strategy. Meantime, top-1 fails to predict all the no-target samples, resulting in 0 \ntacc.~Similar scenarios are observed with analogous top-k strategies, such as the top-5 to top-100 as outlined in \cref{tab:selectingBox}. The pattern remains consistent across these strategies. Therefore, it is evident that \textbf{\textit{employing the top-k strategy does not work for addressing the GREC task}}. Instead, our findings suggest that opting to dynamically determine output bounding boxes based on a confidence threshold proves to be more advantageous. This approach allows the model to dynamically decide the number of bounding boxes required for each specific sample. As shown in \cref{tab:selectingBox}, the ``threshold'' strategy yields the most favorable results in terms of both Pr@(F$_1$=1, IoU$\ge$0.5) and \ntacc~metrics. The choice of the threshold value has a significant impact on the overall performance. To illustrate, adjusting the threshold to 0.9 yields a marked improvement of 14.2\% (37.0\%$\rightarrow$51.2\%) compared to using a threshold of 0.5. This underscores the critical role that threshold selection plays in enhancing performance outcomes. We look forward to more advanced output strategies in the future works from the community.

Notably, the Average Precision (AP)\footnote{Following the primary detection metric of COCO~\cite{lin2014microsoft}, the AP is computed by averaging over different IoU thresholds ranging from 0.50 to 0.95 with a step of 0.05.} metric does not penalize too much the inclusion of numerous redundant bounding boxes characterized by low confidence scores. Consequently, having a greater number of bounding boxes contributes to a higher AP value. However, in the context of REC/GREC, it's imperative to avoid inundating users with an excessive number of redundant bounding boxes when they input expressions targeting specific objects. This outcome is deemed undesirable and falls short of meeting user expectations. Taking this into consideration, it's important to note that the Average Precision (AP) metric doesn't fully capture the performance of REC and GREC.

\begin{figure}[t]
  \begin{center}
     \includegraphics[width=0.996\linewidth]{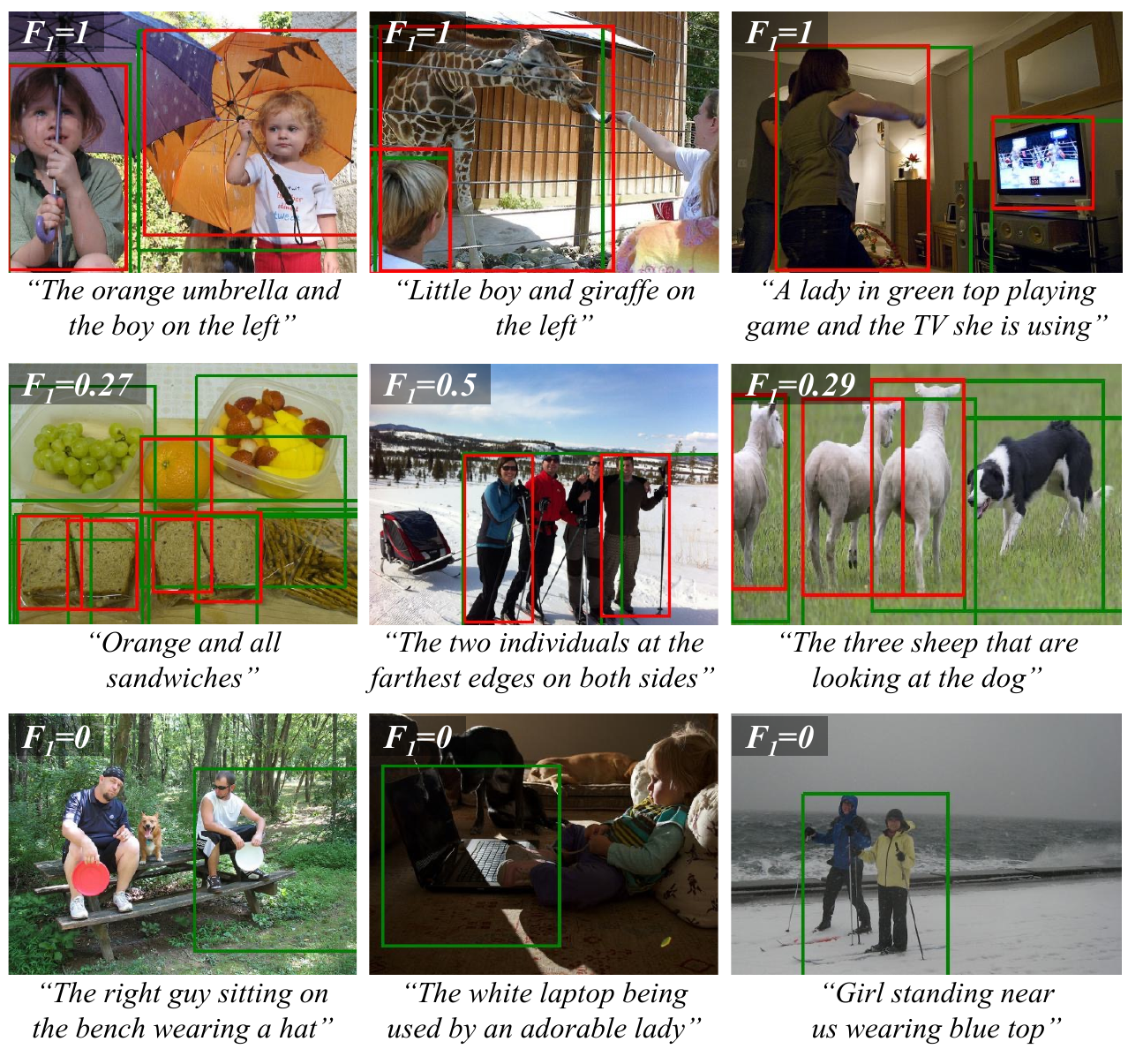}
  \end{center}
 \vspace{-3.6mm}
  \caption{Exemplary results of the modified MDETR~\cite{MDETR} on gRefCOCO dataset. The ground truth is denoted by red bounding boxes, whereas green bounding boxes represent the predictive results. The uppermost row showcases examples of successful outcomes, while the subsequent two rows depict examples of failure cases for multi-target and no-target scenarios, respectively.}
  \label{fig:gRefCOCO-demo}
\end{figure}

\textbf{Qualitative results.} Some qualitative examples of the modified MDETR~\cite{MDETR} on the val set of gRefCOCO are shown in \cref{fig:gRefCOCO-demo}. The ground truth and predictive results are denoted by red bounding boxes and green bounding boxes, respectively. The first row demonstrate examples of successful outcomes, while the subsequent two rows show examples of failure cases for multi-target and no-target scenarios, respectively. The model has the capability to detect salient and unambiguous cues such as \textit{``orange''} and \textit{``on the left''}, but struggles in understanding the shared clues. For example, the first image in the 3rd row has a very deceptive expression of \textit{``The right guy sitting on the bench wearing a hat''}. In the image, a person is situated on the right side and is seated on a bench, but does not have hat. This scenario necessitates the model's capacity to comprehend intricate semantic nuances, enabling it to differentiate between the specific referenced target and the various objects present within the image.

\textbf{GREC benchmark results on gRefCOCO.} In \cref{tab:results_grec}, we report the benchmark results of classic REC methods on gRefCOCO.
These methods have been re-implemented and adjusted to generate numerous bounding boxes. Following this, they employ a threshold-based criterion to determine the final target box(es). As shown in \cref{tab:results_grec}, in contrast to their impressive performance ($\sim$85+\%) on single-target datasets such as RefCOCO, the outcomes of these methods on gRefCOCO exhibit a noticeable decrease. This disparity highlights that these approaches are not effectively tackling the new challenges presented by GREC.

We would like to underscore that any form of training or pre-training must exclude the images from the val, testA, and testB sets of both gRefCOCO and RefCOCO datasets. This precautionary measure is essential to prevent any inadvertent leakage of information.

\section{Conclusion}

This study delves into the constraints associated with the classic Referring Expression Comprehension (REC) task, particularly its incapability to manage multi-target and no-target expressions. Building on this analysis, we introduce a novel benchmark named Generalized Referring Expression Comprehension (GREC), which breaks away from the confines of single-target expressions and enables the inclusion of an arbitrary number of target objects within the expressions. The introduction of GREC fundamentally relaxes the constraints imposed on natural language inputs, expanding the applicability to encompass scenarios involving multiple instances and instances where suitable objects are absent in the image. Additionally, GREC opens up the potential for new applications.

{\small
\bibliographystyle{ieee_fullname}
\bibliography{egbib}

\begin{thebibliography}{10}\itemsep=-1pt

\bibitem{chen2018real}
Xinpeng Chen, Lin Ma, Jingyuan Chen, Zequn Jie, Wei Liu, and Jiebo Luo.
\newblock Real-time referring expression comprehension by single-stage grounding network.
\newblock {\em arXiv preprint arXiv:1812.03426}, 2018.

\bibitem{deng2021transvg}
Jiajun Deng, Zhengyuan Yang, Tianlang Chen, Wengang Zhou, and Houqiang Li.
\newblock Transvg: End-to-end visual grounding with transformers.
\newblock In {\em ICCV}, 2021.

\bibitem{phraseclick}
Henghui Ding, Scott Cohen, Brian Price, and Xudong Jiang.
\newblock Phraseclick: toward achieving flexible interactive segmentation by phrase and click.
\newblock In {\em ECCV}, 2020.

\bibitem{MeViS}
Henghui Ding, Chang Liu, Shuting He, Xudong Jiang, and Chen~Change Loy.
\newblock {MeViS}: A large-scale benchmark for video segmentation with motion expressions.
\newblock In {\em ICCV}, 2023.

\bibitem{MOSE}
Henghui Ding, Chang Liu, Shuting He, Xudong Jiang, Philip~HS Torr, and Song Bai.
\newblock {MOSE}: A new dataset for video object segmentation in complex scenes.
\newblock In {\em ICCV}, 2023.

\bibitem{ding2021vision}
Henghui Ding, Chang Liu, Suchen Wang, and Xudong Jiang.
\newblock Vision-language transformer and query generation for referring segmentation.
\newblock In {\em ICCV}, 2021.

\bibitem{VLTPAMI}
Henghui Ding, Chang Liu, Suchen Wang, and Xudong Jiang.
\newblock {VLT}: Vision-language transformer and query generation for referring segmentation.
\newblock {\em IEEE TPAMI}, 45(6), 2023.

\bibitem{grubinger2006iapr}
Michael Grubinger, Paul Clough, Henning M{\"u}ller, and Thomas Deselaers.
\newblock The iapr tc-12 benchmark: A new evaluation resource for visual information systems.
\newblock In {\em International workshop ontoImage}, volume~2, 2006.

\bibitem{han2023global}
Kunyang Han, Yong Liu, Jun~Hao Liew, Henghui Ding, Yunchao Wei, Jiajun Liu, Yitong Wang, Yansong Tang, Yujiu Yang, Jiashi Feng, et~al.
\newblock Global knowledge calibration for fast open-vocabulary segmentation.
\newblock In {\em ICCV}, 2023.

\bibitem{PADing}
Shuting He, Henghui Ding, and Wei Jiang.
\newblock Primitive generation and semantic-related alignment for universal zero-shot segmentation.
\newblock In {\em CVPR}, 2023.

\bibitem{D2Zero}
Shuting He, Henghui Ding, and Wei Jiang.
\newblock Semantic-promoted debiasing and background disambiguation for zero-shot instance segmentation.
\newblock In {\em CVPR}, 2023.

\bibitem{hong2019learning}
Richang Hong, Daqing Liu, Xiaoyu Mo, Xiangnan He, and Hanwang Zhang.
\newblock Learning to compose and reason with language tree structures for visual grounding.
\newblock {\em IEEE TPAMI}, 44(2), 2022.

\bibitem{hu2017modeling}
Ronghang Hu, Marcus Rohrbach, Jacob Andreas, Trevor Darrell, and Kate Saenko.
\newblock Modeling relationships in referential expressions with compositional modular networks.
\newblock In {\em CVPR}, 2017.

\bibitem{hu2016natural}
Ronghang Hu, Huazhe Xu, Marcus Rohrbach, Jiashi Feng, Kate Saenko, and Trevor Darrell.
\newblock Natural language object retrieval.
\newblock In {\em CVPR}, 2016.

\bibitem{RefCLIP}
Lei Jin, Gen Luo, Yiyi Zhou, Xiaoshuai Sun, Guannan Jiang, Annan Shu, and Rongrong Ji.
\newblock Refclip: A universal teacher for weakly supervised referring expression comprehension.
\newblock In {\em CVPR}, 2023.

\bibitem{MDETR}
Aishwarya Kamath, Mannat Singh, Yann LeCun, Gabriel Synnaeve, Ishan Misra, and Nicolas Carion.
\newblock Mdetr-modulated detection for end-to-end multi-modal understanding.
\newblock In {\em ICCV}, 2021.

\bibitem{kazemzadeh-etal-2014-referitgame}
Sahar Kazemzadeh, Vicente Ordonez, Mark Matten, and Tamara Berg.
\newblock {R}efer{I}t{G}ame: Referring to objects in photographs of natural scenes.
\newblock In {\em {EMNLP}}, 2014.

\bibitem{krishna2017visual}
Ranjay Krishna, Yuke Zhu, Oliver Groth, Justin Johnson, Kenji Hata, Joshua Kravitz, Stephanie Chen, Yannis Kalantidis, Li-Jia Li, David~A Shamma, et~al.
\newblock Visual genome: Connecting language and vision using crowdsourced dense image annotations.
\newblock {\em IJCV}, 123(1), 2017.

\bibitem{li2023transformer}
Xiangtai Li, Henghui Ding, Wenwei Zhang, Haobo Yuan, Jiangmiao Pang, Guangliang Cheng, Kai Chen, Ziwei Liu, and Chen~Change Loy.
\newblock Transformer-based visual segmentation: A survey.
\newblock {\em arXiv:2304.09854}, 2023.

\bibitem{liao2020real}
Yue Liao, Si Liu, Guanbin Li, Fei Wang, Yanjie Chen, Chen Qian, and Bo Li.
\newblock A real-time cross-modality correlation filtering method for referring expression comprehension.
\newblock In {\em CVPR}, 2020.

\bibitem{lin2014microsoft}
Tsung-Yi Lin, Michael Maire, Serge Belongie, James Hays, Pietro Perona, Deva Ramanan, Piotr Doll{\'a}r, and C~Lawrence Zitnick.
\newblock Microsoft coco: Common objects in context.
\newblock In {\em ECCV}, 2014.

\bibitem{GRES}
Chang Liu, Henghui Ding, and Xudong Jiang.
\newblock {GRES}: Generalized referring expression segmentation.
\newblock In {\em CVPR}, 2023.

\bibitem{liu2023multi}
Chang Liu, Henghui Ding, Yulun Zhang, and Xudong Jiang.
\newblock Multi-modal mutual attention and iterative interaction for referring image segmentation.
\newblock {\em IEEE TIP}, 2023.

\bibitem{liu2022instance}
Chang Liu, Xudong Jiang, and Henghui Ding.
\newblock Instance-specific feature propagation for referring segmentation.
\newblock {\em IEEE TMM}, 2022.

\bibitem{liu2019learning}
Daqing Liu, Hanwang Zhang, Feng Wu, and Zheng-Jun Zha.
\newblock Learning to assemble neural module tree networks for visual grounding.
\newblock In {\em ICCV}, 2019.

\bibitem{GroundingDINO}
Shilong Liu, Zhaoyang Zeng, Tianhe Ren, Feng Li, Hao Zhang, Jie Yang, Chunyuan Li, Jianwei Yang, Hang Su, Jun Zhu, et~al.
\newblock Grounding dino: Marrying dino with grounded pre-training for open-set object detection.
\newblock {\em arXiv preprint arXiv:2303.05499}, 2023.

\bibitem{luo2020multi}
Gen Luo, Yiyi Zhou, Xiaoshuai Sun, Liujuan Cao, Chenglin Wu, Cheng Deng, and Rongrong Ji.
\newblock Multi-task collaborative network for joint referring expression comprehension and segmentation.
\newblock In {\em CVPR}, 2020.

\bibitem{mao2016generation}
Junhua Mao, Jonathan Huang, Alexander Toshev, Oana Camburu, Alan~L Yuille, and Kevin Murphy.
\newblock Generation and comprehension of unambiguous object descriptions.
\newblock In {\em CVPR}, 2016.

\bibitem{sun2022proposal}
Mengyang Sun, Wei Suo, Peng Wang, Yanning Zhang, and Qi Wu.
\newblock A proposal-free one-stage framework for referring expression comprehension and generation via dense cross-attention.
\newblock {\em IEEE TMM}, 2022.

\bibitem{wang2019neighbourhood}
Peng Wang, Qi Wu, Jiewei Cao, Chunhua Shen, Lianli Gao, and Anton van~den Hengel.
\newblock Neighbourhood watch: Referring expression comprehension via language-guided graph attention networks.
\newblock In {\em CVPR}, 2019.

\bibitem{wu2020phrasecut}
Chenyun Wu, Zhe Lin, Scott Cohen, Trung Bui, and Subhransu Maji.
\newblock Phrasecut: Language-based image segmentation in the wild.
\newblock In {\em CVPR}, 2020.

\bibitem{wu2023betrayed}
Jianzong Wu, Xiangtai Li, Henghui Ding, Xia Li, Guangliang Cheng, Yunhai Tong, and Chen~Change Loy.
\newblock Betrayed by captions: Joint caption grounding and generation for open vocabulary instance segmentation.
\newblock In {\em ICCV}, 2023.

\bibitem{wu2022towards}
Jianzong Wu, Xiangtai Li, Xia Li, Henghui Ding, Yunhai Tong, and Dacheng Tao.
\newblock Towards robust referring image segmentation.
\newblock {\em arXiv preprint arXiv:2209.09554}, 2022.

\bibitem{wu2023towards}
Jianzong Wu, Xiangtai Li, Shilin Xu~Haobo Yuan, Henghui Ding, Yibo Yang, Xia Li, Jiangning Zhang, Yunhai Tong, Xudong Jiang, Bernard Ghanem, et~al.
\newblock Towards open vocabulary learning: A survey.
\newblock {\em arXiv:2306.15880}, 2023.

\bibitem{UNINEXT}
Bin Yan, Yi Jiang, Jiannan Wu, Dong Wang, Zehuan Yuan, Ping Luo, and Huchuan Lu.
\newblock Universal instance perception as object discovery and retrieval.
\newblock In {\em CVPR}, 2023.

\bibitem{yang2020improving}
Zhengyuan Yang, Tianlang Chen, Liwei Wang, and Jiebo Luo.
\newblock Improving one-stage visual grounding by recursive sub-query construction.
\newblock In {\em ECCV}, 2020.

\bibitem{yang2019fast}
Zhengyuan Yang, Boqing Gong, Liwei Wang, Wenbing Huang, Dong Yu, and Jiebo Luo.
\newblock A fast and accurate one-stage approach to visual grounding.
\newblock In {\em ICCV}, 2019.

\bibitem{yu2016modeling}
Licheng Yu, Patrick Poirson, Shan Yang, Alexander~C Berg, and Tamara~L Berg.
\newblock Modeling context in referring expressions.
\newblock In {\em ECCV}, 2016.

\bibitem{zhanghui2021}
Hui Zhang and Henghui Ding.
\newblock Prototypical matching and open set rejection for zero-shot semantic segmentation.
\newblock In {\em ICCV}, 2021.

\bibitem{zhang2018grounding}
Hanwang Zhang, Yulei Niu, and Shih-Fu Chang.
\newblock Grounding referring expressions in images by variational context.
\newblock In {\em CVPR}, 2018.

\bibitem{zhuang2018parallel}
Bohan Zhuang, Qi Wu, Chunhua Shen, Ian Reid, and Anton Van Den~Hengel.
\newblock Parallel attention: A unified framework for visual object discovery through dialogs and queries.
\newblock In {\em CVPR}, 2018.

\end{thebibliography}
}

\end{document}